\documentclass[conference]{IEEEtran}
\IEEEoverridecommandlockouts
\usepackage{CJKutf8}
\usepackage{cite}
\usepackage{amsmath,amssymb,amsfonts}
\usepackage{algorithmic}
\usepackage{graphicx}
\usepackage{textcomp}
\usepackage{xcolor}
\usepackage{multirow}
\usepackage{booktabs}
\usepackage{url}
\usepackage{float}
\usepackage{subcaption}
\def\BibTeX{{\rm B\kern-.05em{\sc i\kern-.025em b}\kern-.08em
    T\kern-.1667em\lower.7ex\hbox{E}\kern-.125emX}}
\begin{document}

\title{Leveraging Zipformer Model for Effective Language Identification in Code-Switched Child-Directed Speech\\
}

\author{
\IEEEauthorblockN{
Lavanya Shankar,
Leibny Paola Garcia Perera
}
\IEEEauthorblockA{
Johns Hopkins University, Baltimore, USA \quad
}
\IEEEauthorblockA{
\texttt{\{ls1, lgarci27\}@jhu.edu}
}
}

\maketitle

\begin{abstract}
Code-switching and language identification in child-directed scenarios present significant challenges, particularly in bilingual environments. This paper addresses this challenge by using \texttt{Zipformer} to handle the nuances of speech which contains two imbalanced languages -- Mandarin and English -- in an utterance. This work demonstrates that the internal layers of the \texttt{Zipformer} effectively encode the language characteristics, which can be leveraged in language identification. We present the selection methodology of the inner layers to extract the embeddings and make a comparison with different back-ends. Our analysis shows that \texttt{Zipformer} is robust across these backends. Our approach effectively handles imbalanced data, achieving a Balanced Accuracy (BAC) of 81.89\%, a 15.47\% improvement over the language identification baseline. These findings highlight the potential of the transformer encoder architecture model in real scenarios.

\end{abstract}


\section{Introduction}

Language identification (LID) is a fundamental task in speech processing that enables systems to determine the language spoken in an audio segment. Although LID has been extensively studied for high-resource adult speech, low-resource child-directed speech (CDS) remains a relatively unexplored domain. CDS involves speech interactions between caregivers and children, which differ significantly from adult speech ~\cite{jones2023characteristics, cristia2019segmentability} in terms of lexical choices, higher pitch, slower speech rate, and increased prosodic variation. Moreover, LID in CDS is particularly challenging due to the low-resource nature of the data. This is further complicated by the frequent presence of code-switching, in which multiple languages alternate within a conversation.

One such dataset is the MERLIon CCS corpus \cite{chua2023merlion}, a Singaporean bilingual speech data featuring English-Mandarin code-switching. 
In its raw form, this dataset presents a significant language imbalance, with a higher number of English utterances compared to Mandarin. Even with ground-truth language annotations, traditional LID \cite{jung2024language} models struggle with such imbalances, often biasing their predictions toward the majority language.

To address these challenges, we explore the use of Zipformer \cite{zipformer}, a transformer-based model originally developed for automatic speech recognition (ASR) that features a U-Net-like encoder structure. Unlike standard transformers, Zipformer processes speech at multiple frame rates by downsampling and upsampling the sequence in its middle stacks, enabling hierarchical modeling of both local and global dependencies. This structure is particularly effective for capturing the slow speech rates and exaggerated prosody characteristic of CDS. To our knowledge, this work presents the first comprehensive adaptation and evaluation of Zipformer for LID in code-switched CDS. Recent research suggests that ASR models encode linguistic features at intermediate layers ~\cite{hussein2024enhancing, shen2023generative, belinkov2017hidden}, making them useful for downstream tasks such as LID. Our approach involves extracting embeddings from these intermediate layers and training a set of classifiers to distinguish between languages. We perform a comprehensive layer-wise analysis of Zipformer embeddings to identify the most informative representations for LID. In addition, we conduct an extensive comparison of four backend classifiers. We also include a baseline using MFCC embeddings with three machine learning models for comparison. With \texttt{Zipformer}, we develop a more robust LID system capable of handling low-resource, imbalanced, and code-switched speech data.

\section{Related Work}
Significant research has been conducted in the field of LID, primarily focusing on adult speech. However, LID for child-directed speech (CDS) remains less explored, particularly under low-resource and code-switched settings. The availability of the MERLIon CCS corpus~\cite{chua2023merlion} has recently helped to address this gap.

Several recent studies have used the MERLIon CCS corpus~\cite{chua2023merlion} as their primary dataset. ~\cite{stella} employed a stacked CNN-GRU model with a transfer learning approach, incorporating ASR as a secondary task to capture speech patterns and language characteristics. Similarly,~\cite{wang24f_interspeech} proposed a phonetics-based LID and diarization framework for CDS, using a convolutional layer followed by a transformer encoder to model phoneme context and temporal dependencies. Their approach extends the PHO-LID model~\cite{liu2021end} by integrating acoustic and phonotactic information.~\cite{gupta2024spoken} presented a two-stage end-to-end model featuring a convolutional encoder with Squeeze-and-Excitation blocks and an attentive temporal pooling decoder. Their design prioritizes efficiency, employing fewer parameters than large-scale pre-trained speech models. Furthermore, ~\cite{shahin2023improving} used a self-supervised learning (SSL) approach with a wav2vec2 model and performed binary classification on code-switched data.

Although these works advanced LID for CDS, most did not provide in-depth analyses of how internal layers contribute to LID. While transformer-based architectures, including Conformer models, have started to appear, such as in the MERLIon CCS Challenge baseline system~\cite{chua2023merlion} and in the transformer encoder used by~\cite{wang24f_interspeech}, comprehensive interpretability studies remain limited. Furthermore, most approaches evaluated performance using only a single back-end classifier, which restricts understanding of how classifier choice influences outcomes in CDS-specific settings.

In contrast, adult speech LID has seen extensive work using transformer-based and SSL architectures. For example, ~\cite{bartley2022accidental} demonstrated that the Conformer architecture, when used within a multilingual self-supervised pre-training framework, can effectively encode language identification information in its lower layers.~\cite{gurusamy2023transformer} provided a comprehensive comparison of transformer models, including BERT and RoBERTa, showing that these architectures outperform traditional methods and have established a new benchmark for adult speech LID in NLP.

Although these studies do not focus on CDS, they illustrate the effectiveness of transformer and attention-based methods in handling noisy, real-world environments. However, they also lack in-depth layer-wise analysis and evaluation across multiple classifiers.

Despite these advances, there remains a clear need for comprehensive studies that both interpret the internal representations of transformer-based models and systematically evaluate the impact of different backend classifiers for LID. Addressing these gaps is essential for developing robust and generalizable LID systems that are tailored to the unique characteristics of CDS.

\section{Methodology}
\begin{figure} [t]
\footnotesize
    \centering
    \includegraphics[width=0.9\linewidth]{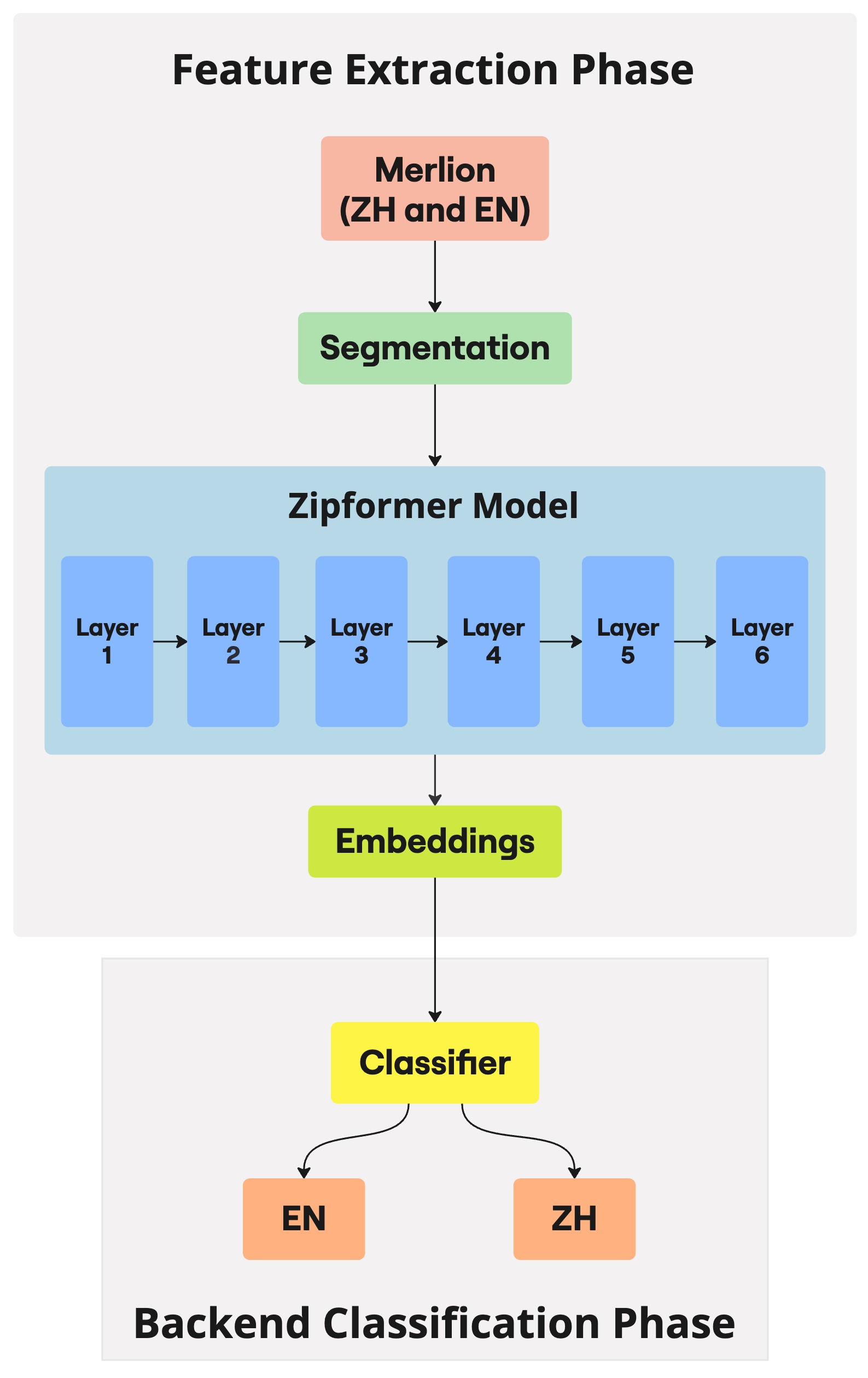}
    \caption{Model Architecture: The MERLIon dataset is processed using a pre-trained Zipformer model to extract embeddings from all six layers. These embeddings are then utilized for classification using various models.}
    \label{fig:enter-label1}
\end{figure}

Our research extends the use of the \texttt{Zipformer} model to CDS, addressing challenges such as data imbalance. By integrating transformer-based feature extraction with different backend classifiers, our study contributes to advancing LID performance.
Figure \ref{fig:enter-label1} shows the complete methodology pipeline.\footnote{The code is publicly available at: 

\noindent https://anonymous.4open.science/r/languageIdentification-5513/README.md}

\subsection{Front-End: Feature Extraction}

In the feature extraction phase, the system extracts embeddings using the \texttt{Zipformer} model. These embeddings are then used for downstream tasks, including LID. 

\begin{figure}[h]
    \centering
    \includegraphics[width=1\linewidth]{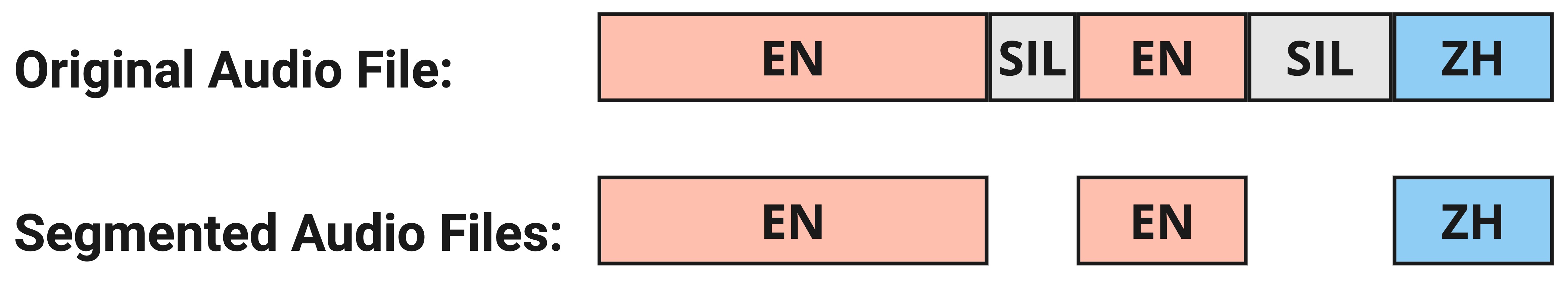}
    \caption{Segmentation of audio data based on ground-truth timestamps.}
    \label{fig:segmentation}
\end{figure}

The system segments the data based on the provided ground-truth timestamps, as shown in Figure \ref{fig:segmentation}. The ground-truth provides a controlled environment to test the reliability of the LID system.

Once segmented, the audio files are processed through the pre-trained \texttt{Zipformer} model, which consists of six layers, each capturing different aspects of the audio data. The embedding sizes for each layer are summarized in Table \ref{tab:embedding_sizes}. These embeddings provide hierarchical representations of the audio, with each layer contributing distinct feature dimensions that enhance the model's ability to understand linguistic nuances.
\begin{table}[htpb]
\footnotesize
  \caption{Embedding sizes for each layer of the Zipformer model.}
  \label{tab:embedding_sizes}
  \centering
  \begin{tabular}{c c c}
    \toprule
    \textbf{Layer} & \textbf{Embedding Size} \\
    \midrule
    Layer 1 & 192 \\
    Layer 2 & 256 \\
    Layer 3 & 384 \\
    Layer 4 & 512 \\
    Layer 5 & 384 \\
    Layer 6 & 256 \\
    \bottomrule
  \end{tabular}
\end{table}

We extract embeddings from each layer of the \texttt{Zipformer} model and evaluate their performance in a classification task. Specifically, we analyze the performance of the different layers{\it, i.e.,} the contribution to classification accuracy. In addition, we identify the layers that produce the most discriminative features for LID.


\subsection{Backend: Classification Phase}

In the backend classification phase, we utilize embeddings extracted from all six layers of the \texttt{Zipformer} model. We experimented with four different architectures to test the model: LSTM, BiLSTM, hybrid CNN-BiLSTM, and Transformer.

\begin{itemize}
    \item \textbf{LSTM Model:} Consists of an LSTM layer followed by a fully connected classification layer.
    \item \textbf{BiLSTM Model:} Employs a bidirectional LSTM layer followed by a fully connected classification layer.
    \item \textbf{Hybrid CNN-BiLSTM Model:} Combines convolutional layers with BiLSTM layers, followed by a fully connected classification layer.
    \item \textbf{Transformer Model:} Incorporates self-attention mechanisms and positional encodings to capture sequential dependencies in the data.
\end{itemize}

\section{Experiments}
\subsection{Data Analysis}
The MERLIon dataset consists of a list of audio files 
containing a total of 28.61 hours of audio, as described in Table \ref{tab:merlion_overview}

\begin{table}[ht]
\footnotesize
\caption{Overview of the MERLIon Speech Dataset.}
\label{tab:merlion_overview}
\centering
\begin{tabular}{lrr}
\toprule
\textbf{Category} & \textbf{Count} & \textbf{Percentage (\%)} \\
\midrule
English & 40,287 & 66.8 \\
Mandarin & 9,983 & 16.5 \\
Non-speech & 10,090 & 16.7 \\
\midrule
\multicolumn{3}{l}{\textit{Total audio duration: 28.61 hours}} \\
\bottomrule
\end{tabular}
\end{table}

 We can observe that 66.8\% of the data is in English, 16.5\% is in Mandarin, and the remaining portion consists of non-speech data. The data is imbalanced, which is common in code-switching datasets. 

\subsection{Data Preparation}

We built a pipeline where audio files are loaded and resampled.\footnote{The pre-trained \texttt{Zipformer} model requires audio files in a 16 kHz format, so we preprocess the dataset accordingly.} We used \texttt{Lhotse} \footnote{\url{https://lhotse.readthedocs.io/en/v1.25.0/index.html}} and \texttt{PyTorch} \footnote{\url{https://pytorch.org/}} for this purpose. We remove non-speech and segment the data based on the provided timestamps. 
Each audio segment is annotated with metadata, including language tags, durations, and unique identifiers.

We randomly selected  70\% of the segments for training, 15\% for validation, and 15\% for testing. This split allows us to evaluate the model's generalization capabilities on unseen data, ensuring that the model does not simply memorize the training examples. 

\subsection{Pre-trained Model}

 We used the pre-trained \texttt{Zipformer} model available on Hugging Face \footnote{\url{https://huggingface.co/zrjin/icefall-asr-zipformer-multi-zh-en-2023-11-22/}} for our experiments. This model's architecture and training on both English and Chinese datasets make it particularly effective for our code-switching task. 

The training sets used in the pretrained \texttt{Zipformer} model include:

\begin{itemize}
    \item \textbf{LibriSpeech (English)}: A widely used dataset for English speech recognition, comprising 960 hours of audio.~\cite{librispeech}
    \item \textbf{AiShell-2 (Chinese)}: A comprehensive dataset for Mandarin Chinese speech recognition, containing 1,000 hours of audio.~\cite{aishell2}
    \item \textbf{TAL-CSASR (Code-Switching, Chinese and English)}: This dataset focuses on code switching between Chinese and English, providing 587 hours of audio data.~\cite{talcs}
\end{itemize}

\subsection{Training}



All the precomputed segments in the MERLIon dataset were passed through the \texttt{Zipformer}, generating embeddings for each layer.
Each embedding was associated with its corresponding ground truth label. 

We experimented with four architectures: LSTM, BiLSTM, hybrid CNN-BiLSTM, and Transformer. The LSTM model has one LSTM layer to capture temporal dependencies followed by a fully connected classification layer. The BiLSTM model used two LSTM layers to capture both forward and backward temporal dependencies, followed by a fully connected layer. In the hybrid CNN-BiLSTM model, we added three Conv1D layers to capture local patterns, each followed by BatchNorm and Dropout for stability. We then incorporated two BiLSTM layers to learn long-range dependencies, followed by fully connected layers and a softmax output layer for classification. The Transformer model consists of an embedding layer for input representation, positional encoding for sequence information, transformer encoder layers for attention-based processing, and a fully connected layer for final predictions.

Training was performed with a batch size of 32 over 20 epochs. The Adam optimizer was employed with a learning rate of 0.001, and the cross-entropy loss function was used to evaluate the model's performance. To prevent overfitting, we monitored validation loss and accuracy after each epoch and implemented early stopping if the validation performance did not improve over a specified number of epochs.


\subsection{Evaluation}
Each model was evaluated on the test set to assess its performance. Various metrics, including accuracy (ACC), balanced accuracy (BAC), F1 score, and Equal Error Rate (EER), were used. By analyzing these metrics, we aim to determine the impact of embeddings from different layers of the \texttt{Zipformer} model on LID accuracy.

\section{Results}


In our experiments, we evaluated three simple baseline models using MFCC:  XGBoost, Random Forest, and Deep Neural Network (DNN). These models were selected to provide a comprehensive comparison of traditional machine learning approaches for speech processing tasks. XGBoost is a gradient boosting algorithm, while Random Forest is an ensemble method. Default parameters were used for both. The DNN model consists of an input layer, six hidden layers, and a softmax output layer for classification. It uses ReLU activation functions for the hidden layers, categorical cross-entropy as the loss function, and the Adam optimizer. The model was trained for 20 epochs. Among these models, XGBoost slightly outperformed the others, as shown in Table \ref{tab:ML}.

\begin{table}[ht]
  \caption{Performance Comparison of XGBoost, Random Forest, and Deep Neural Network on MFCC Embeddings.}
  \label{tab:ML}
  \centering
  \begin{tabular}{c c c c c}
    \toprule
    \textbf{Model} & \textbf{ACC (\%)} & \textbf{BAC (\%)} & \textbf{F1} & \textbf{EER} \\
    \midrule
    XGBoost & 82.36 & 50.38 & 0.75 & 0.496 \\
    Random Forest & 82.26 & 50.0 & 0.74 & 0.503 \\
    DNN & 81.94 & 50.5 & 0.75 & 0.496 \\
    \bottomrule
  \end{tabular}
\end{table}

For the \texttt{Zipformer}, we evaluated a Long Short-Term Memory (LSTM) model across six layers while keeping the number of epochs constant at 15. Layer 3 demonstrated the highest accuracy, as summarized in Table \ref{tab:lstm_results}.

\begin{table}[ht]
\footnotesize
  \caption{Layer-wise Performance of LSTM Classifier with Fixed 15 Epochs.}
  \label{tab:lstm_results}
  \centering
  \begin{tabular}{c c c c c}
    \toprule
    \textbf{Layer} & \textbf{ACC (\%)} & \textbf{BAC (\%)} & \textbf{F1} & \textbf{EER} \\
    \midrule
    1 & 80.67 & 50.06 & 0.613 & 0.499 \\
    2 & 88.27 & 77.78 & 0.669 & 0.222 \\
    3 & 90.64 & 81.29 & 0.731 & 0.188 \\
    4 & 86.48 & 70.21 & 0.562 & 0.298 \\
    5  & 88.93 & 76.36 & 0.664 & 0.236 \\
    6 & 89.99 & 58.48 & 0.697 & 0.215 \\
    \bottomrule
  \end{tabular}
\end{table}

For the BiLSTM model, we followed a similar approach, but with additional techniques to prevent overfitting, such as early stopping and mixed-precision training. As observed from Table \ref{tab:bilstm_results}, the number of epochs varied for each layer due to early stopping. The model ran for more epochs in Layer 3, achieving an accuracy of 90.71\%.

\begin{table}[ht]
\footnotesize
\caption{Layer-wise Performance of BiLSTM Classifier with Adaptive Epochs.}
\label{tab:bilstm_results}
\centering
\begin{tabular}{cccccc}
\toprule
\textbf{Layer} & \textbf{Epoch} & \textbf{ACC} & \textbf{BAC} & \textbf{F1} & \textbf{EER} \\
& & \textbf{(\%)} & \textbf{(\%)} & & \\
\midrule
1 & 11 & 80.10 & 50.26 & 0.714 & 0.446 \\
2 & 13 & 81.45 & 50.15 & 0.732 & 0.526 \\
\textbf{3} & \textbf{19} & \textbf{90.71} & \textbf{81.89} & \textbf{0.904} & \textbf{0.157} \\
4 & 12 & 84.49 & 66.51 & 0.827 & 0.364 \\
5 & 8 & 81.33 & 51.57 & 0.737 & 0.493 \\
6 & 10 & 80.65 & 50.56 & 0.723 & 0.487 \\
\bottomrule
\end{tabular}
\end{table}

\begin{figure}
    \centering
    \includegraphics[width=1\linewidth]{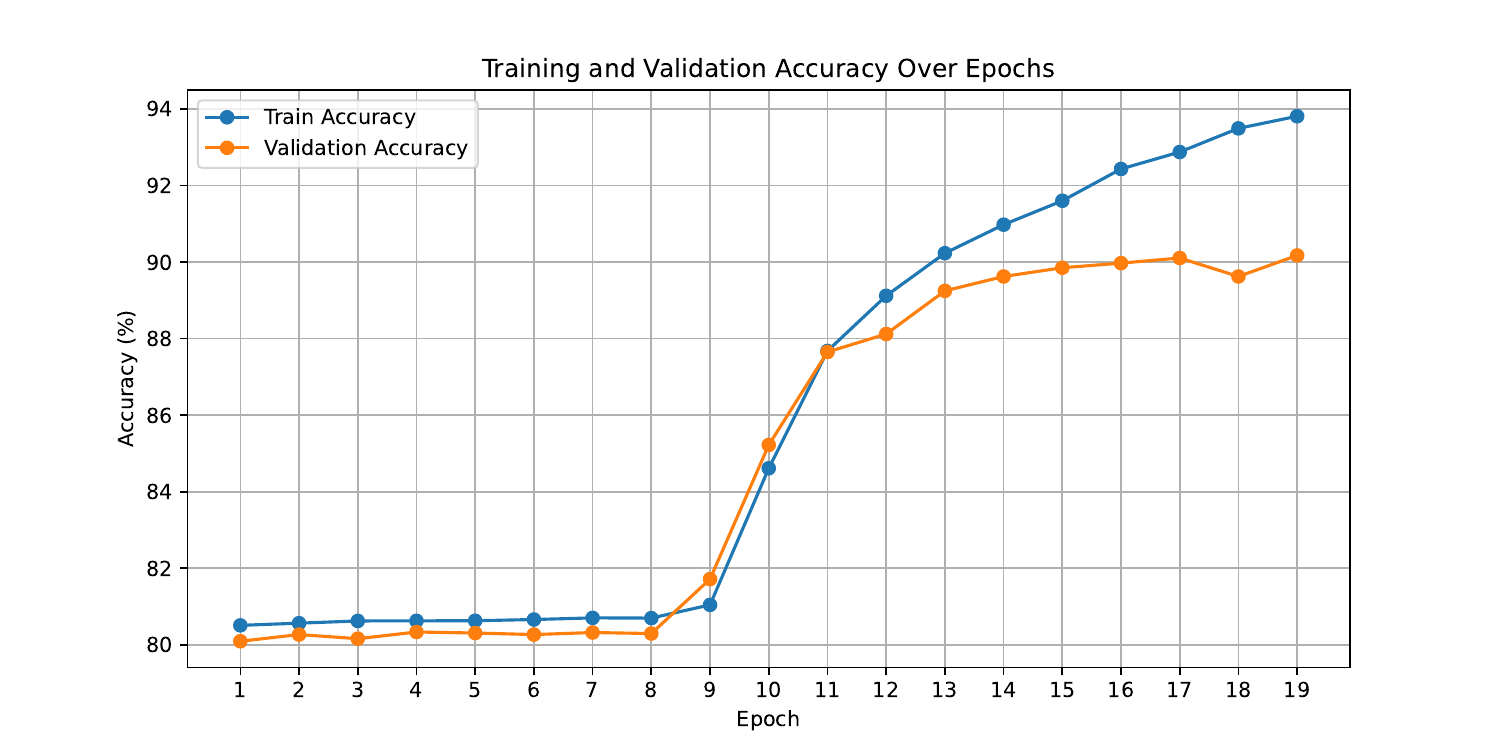}
    \caption{Accuracy Growth of BiLSTM Classifier in Layer 3.}
    \label{fig:plt}
\end{figure}

These results indicate that the BiLSTM model performed best in Layer 3, with a high F1 score, suggesting that it effectively balances precision and recall. Figure \ref{fig:plt} visualizes the progression of the accuracy of BiLSTM in Layer 3, demonstrating steady improvement without overfitting.

We also experimented with complex models such as the CNN-BiLSTM hybrid model and transformers. We only performed classification on layer 3 as it gave better results in previous setups. As shown in Table \ref{tab:complex}, the results were not satisfactory. We deduced that the BiLSTM-CNN model struggled to extract features effectively due to the high dimensionality of the embeddings. For transformers, we believe they perform better with raw input data.

\begin{table}[ht]
\footnotesize
  \caption{Result of Hybrid BiLSTM-CNN model only for Layer 3.}
  \label{tab:complex}
  \centering
  \begin{tabular}{c c c c c c}
    \toprule
\textbf{Model} & \textbf{Epoch} & \textbf{ACC} & \textbf{BAC} & \textbf{F1} & \textbf{EER} \\
& & \textbf{(\%)} & \textbf{(\%)} & & \\
    \midrule
    BiLSTM-CNN & 6 & 80.11 & 50.00 & 0.718 & 0.500 \\
    Transformers & 9 & 80.88 & 50.00 & 0.724 & 0.510 \\
   
    \bottomrule
  \end{tabular}
\end{table}

The Table \ref{tab:lid_performance} presents the results from the paper \cite{wang24f_interspeech}, which focused on LID using the MERLIon dataset. They used a different split compared to ours and incorporated an additional dataset, SEAME \cite{lyu2010seame}, for training. As shown, the highest accuracy achieved was 95.57\%, surpassing our accuracy of 90.71\%. However, we achieved a higher balanced accuracy of 81.89\%, compared to their 66.42\%. Additionally, our F1 score is 90.4\%, significantly higher than their 18.99\%. While their EER rate is better at 4.43\%, ours is 15.7\%.

\begin{table}[ht]
\footnotesize
  \caption{LID Performances from ~\cite{wang24f_interspeech} on MERLIon dataset.}
  \label{tab:lid_performance}
  \centering
  \begin{tabular}{c c c c c}
    \toprule
    \textbf{Model} & \textbf{Acc (\%)} & \textbf{BAC (\%)} & \textbf{F1 (\%)} & \textbf{EER (\%)} \\
    \midrule
    PHO-LID & 95.48 & 65.39 & \textbf{18.66} & 4.524 \\
    conv & 88.41 & \textbf{66.42} & 10.23 & 11.590 \\
    pho emb & 95.43 & 60.75 & 14.19 & 4.571 \\
    conv+pho emb & \textbf{95.57} & 57.75 & 11.34 & \textbf{4.430} \\
    \bottomrule
    \textbf{Zipformer}  & \textbf{90.71} & \textbf{81.89} & \textbf{90.4} & \textbf{15.7} \\
    \bottomrule
  \end{tabular}
\end{table}

\section{Discussion}

\subsection{Zipformer ASR Output}
As part of our evaluation of the \texttt{Zipformer} model's performance in language identification, we examined the output generated during the embedding extraction process. Below are examples of the automatic speech recognition (ASR) output for segments of audio processed by the model:
\begin{CJK*}{UTF8}{gbsn}  
\begin{verbatim}
English Segment:
["THERE'S", 'A', 'RAIN', 'AND', 'WATER']

Mandarin Segement:
['那', '么', '在', '哪', '里']
\end{verbatim}
\end{CJK*}  

These outputs demonstrate the model's ability to transcribe both English and Mandarin segments accurately, showcasing its potential for effective language identification in code-switching scenarios.

\subsection{Embedding Analysis}
To investigate the presence of distinct grouping within the data, we performed clustering on the embeddings extracted from the \texttt{Zipformer}. Specifically, we aimed to determine whether the embeddings could be effectively grouped into two clusters, corresponding to the two language classes: English and Mandarin.

\begin{figure} 
    \centering
    \includegraphics[width=0.8\linewidth]{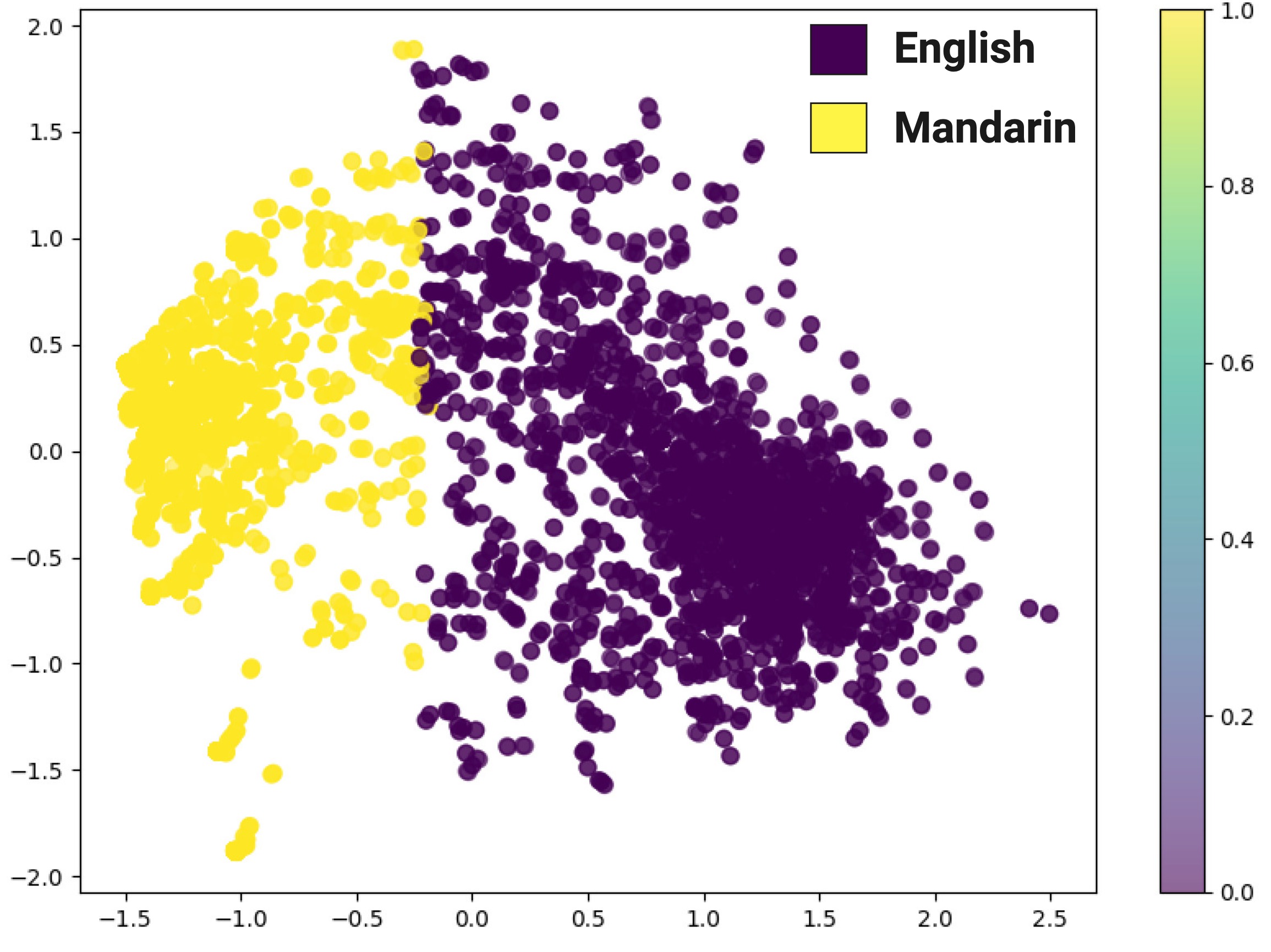}
    \caption{PCA-based clustering showing two clusters for embeddings from Layer 3.}
    \label{fig:PCA}
\end{figure}

For clustering, we used K-Means, ~\cite{hartigan1979algorithm} and then, to visualize the results, we employed both Principal Component Analysis (PCA)~\cite{abdi2010principal} and t-Distributed Stochastic Neighbor Embedding (t-SNE) ~\cite{van2008visualizing}. 
As shown in Figure \ref{fig:PCA}, we could observe two clusters in the embeddings. The embeddings processed through PCA 
resulting in a silhouette score of 0.473. This result suggests a moderate clustering quality due to the complexity of the data.

\begin{figure}
    \centering
    \includegraphics[width=0.8\linewidth]{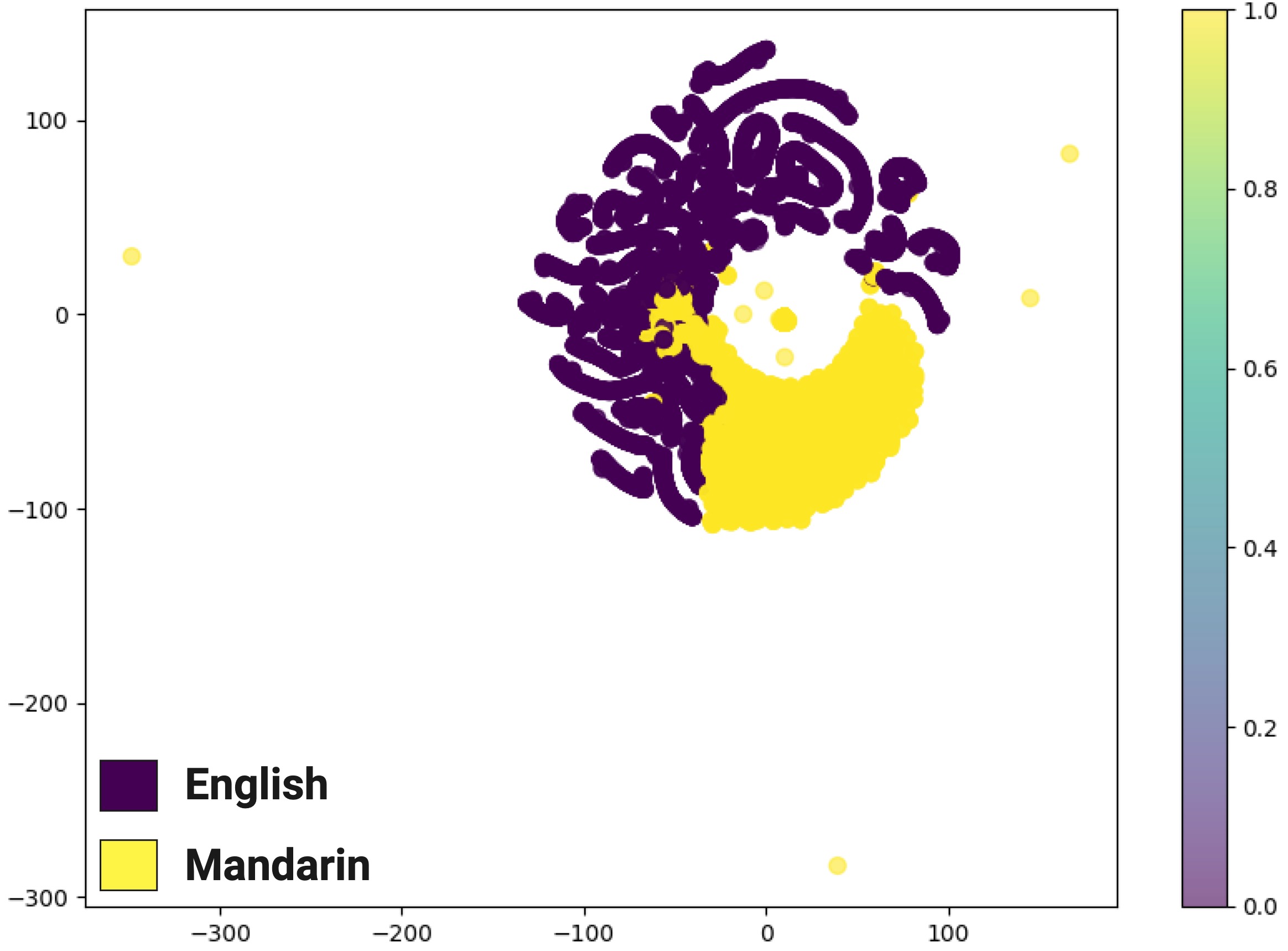}
    \caption{t-SNE based clustering showing two clusters for embeddings from Layer 3.}
    \label{fig:tsne}
\end{figure}

However, to capture more complex, non-linear relationships in the data, we also applied t-SNE, which is particularly well-suited for visualizing high-dimensional data in a lower-dimensional space. As shown in Figure \ref{fig:tsne}, we could observe 2 clusters with a silhouette score of 0.485. Once again, this result suggests a moderate clustering quality due to the complexity of the data.

The silhouette scores from t-SNE are slightly better than those from PCA. In the visualizations from both PCA and t-SNE, we can observe two clusters. This dual approach allowed us to assess whether the embeddings effectively captured the linguistic characteristics of the two languages, further validating the performance of our LID models.

\section{Conclusion}
In this work, we explored LID using embeddings extracted from the \texttt{Zipformer} model. Our experiments involved training and evaluating both LSTM and BiLSTM models to classify embeddings into two language classes: English and Mandarin. The results demonstrated strong performance, with the BiLSTM model achieving the highest accuracy and F1 score.

Additionally, we performed clustering on the embeddings using PCA and t-SNE. Both methods revealed two distinct clusters, corresponding to the two languages, further validating the effectiveness of the \texttt{Zipformer} embeddings for LID tasks. 

For future work, we plan to explore further refinements to the models, such as incorporating additional language pairs and exploring more advanced clustering techniques to improve classification accuracy.
\clearpage
\bibliographystyle{IEEEtran}
\bibliography{mybib}

\end{document}